\documentclass[letterpaper, 10 pt, conference]{ieeeconf}
\IEEEoverridecommandlockouts

\usepackage[utf8]{inputenc}

\usepackage{mathbbol}

\usepackage{esvect}

\usepackage[percent]{overpic}

\usepackage{tabu}

\usepackage{multirow}

\usepackage[overload]{empheq}

\usepackage{bm}

\usepackage{xcolor}
\definecolor{dg}{rgb}{0.1, 0.6, 0.2}       
\definecolor{b}{rgb}{0.0, 0.0, 1}          

\usepackage{epsfig}
\usepackage{float}
\usepackage{color}
\usepackage{url}

\usepackage{algorithmic}
\usepackage[ruled,linesnumbered,vlined]{algorithm2e}

\usepackage{amssymb, amsfonts}
\usepackage{amsmath}
\usepackage{mathrsfs}

\usepackage{amsthm}
\usepackage{mathtools}

\let\labelindent\relax
\usepackage{enumitem}       
\setenumerate[enumerate]{align=left}

\usepackage{graphicx}
\usepackage{subcaption}
\usepackage{caption}

\usepackage{cite}

\makeatletter
\let\NAT@parse\undefined
\makeatother
\usepackage{hyperref} 
\usepackage{footmisc}   

\usepackage{balance}

\newcommand{\floor}[1]{\left\lfloor #1 \right\rfloor}

\newcommand{\norm}[1]{\left\lVert#1\right\rVert}

\makeatletter
\newlength\tmp@\newlength\t@mp
\newcommand{\comp}[3]
  {\mathop{ \settowidth\tmp@{$\displaystyle\mathop{#1}^{#3}_{#2}$}
  \hbox to \tmp@{\hss \settowidth\t@mp{$\displaystyle #1$}\setlength\t@mp{.45\t@mp}
  $\displaystyle\mathop{#1}^{\hspace\t@mp #3}_{\hspace{-\t@mp}#2}$
  \hss} }}
\makeatother

\newcommand{\Int}[2]
{\int_{#1}^{#2}}

\hypersetup{
    colorlinks=true,
    linkcolor=blue,
    filecolor=blue,      
    urlcolor=blue,
    citecolor=blue,
}

\def\a{\alpha}
\def\b{\beta}

\def\D{\Delta}
\def\S{\Sigma}
\def\G{\Gamma}

\def\R{\mathbb{R}}

\def\l{\left}
\def\r{\right}

\def\quat{\mathbf{q}}

\def\pos{\mathbf{p}}
\def\vel{\mathbf{v}}
\def\accel{\mathbf{a}}

\newcommand{\sksym}[1]{\floor{#1}_{\times}}

\newcommand{\vbf}[1]{{\bm{\mathbf{#1}}}}

\def\bias{\mathbf{b}}

\def\rot{\mathbf{R}}
\def\tf{\mathbf{T}}

\def\SO{\mathrm{SO(3)}}

\def\Exp{\mathrm{Exp}}
\def\Log{\mathrm{Log}}

\def\Scov{\textbf{C}}
\def\Smean{\textbf{S}}
\def\doteq{\leftarrow}

\def\X{\mathcal{X}}

\def\I{\mathcal{I}}
\def\V{\mathcal{V}}

\def\L{\mathcal{L}}
\def\P{\mathcal{P}}

\def\T{\mathcal{T}}

\def\angvel{\bm{\omega}}
\def\accel{\mathbf{a}}

\def\f{\vbf{f}}
\def\n{\vbf{n}}

\newcommand{\ul}[1]{\underline{#1}}
\newcommand{\tb}[1]{\textbf{#1}}

\begin{document}

\title{\bf SLICT: Multi-input Multi-scale Surfel-Based Lidar-Inertial\\ Continuous-Time Odometry and Mapping }

\author{
      Thien-Minh Nguyen$^{1*}$,
      Daniel Duberg$^1$,
      Patric Jensfelt$^1$,\\
      Shenghai Yuan$^2$,
      Lihua Xie$^2$, \IEEEmembership{Fellow,~IEEE}
\thanks{$^1$ Division of Robotics, Perception and Learning, KTH Royal Institute of Technology, Stockholm, 11428,
(e-mail: tmng@kth.se, dduberg@kth.se, patric@kth.se).
}
\thanks{$^2$ School of Electrical and Electronic Engineering, Nanyang Technological University, Singapore 639798, 50 Nanyang Avenue. (e-mail: shyuan@ntu.edu.sg, elhxie@ntu.edu.sg)}
\thanks{
This work was partially supported by the \textit{Wallenberg AI, Autonomous Systems and Software Program (WASP)}, funded by the Knut and Alice Wallenberg Foundation, under the Wallenberg-NTU Presidential Postdoctoral Fellowship Program.
}
\thanks{$^*$ Corresponding author.}
}

\maketitle

\begin{abstract}

While feature association to a \textit{global map} has significant benefits, to keep the computations from growing exponentially, most lidar-based odometry and mapping methods opt to associate features with \textit{local maps} at one voxel scale.
Taking advantage of the fact that \textit{surfels} (surface elements) at different voxel scales can be organized in a tree-like structure,
we propose an octree-based global map of \textit{multi-scale} surfels that can be updated incrementally. This alleviates the need for recalculating, for example, a k-d tree of the whole map repeatedly.
The system can also take input from a single or a number of sensors, reinforcing the robustness in degenerate cases.
We also propose a point-to-surfel (PTS) association scheme, continuous-time optimization on PTS and IMU preintegration factors, along with loop closure and bundle adjustment, making a complete framework for Lidar-Inertial continuous-time odometry and mapping.
Experiments on public and in-house datasets demonstrate the advantages of our system compared to other state-of-the-art methods. To benefit the community, we release the source code and dataset at \url{https://github.com/brytsknguyen/slict}.

\end{abstract}

\IEEEpeerreviewmaketitle

\section{Introduction}

Thanks to reduced cost and weight, lidar-based navigation systems have made many significant progresses for autonomous systems in recent years, motivating many new advanced algorithms.
In this paper, we are concerned with two main aspects of a lidar-inertial odometry and mapping (LIOM) system, namely feature extraction-association, and mapping.

For feature extraction-association (also termed the \textit{frontend} \cite{cadena2016past}), existing methods can be grouped into three main categories. In the first one, planar and corner features are extracted based on the smoothness value, then associated with k-nearest neighbours to construct the corresponding cost factors. This method was proposed in \cite{zhang2014loam} and is still widely adopted in many recent works \cite{shan2020liosam, nguyen2021miliom, chen2021low, wang2021floam, lv2021clins}. {One issue of this method is that the smoothness calculation is specialized for the lidar scans with horizontally separated rings, in environments with man-made structure}. To overcome this issue, in FAST-LIO \cite{xu2021fast, xu2022fast}, Xu et al propose the \textit{direct method},
where each lidar point is directly associated with a neighborhood in the map. The method can work well with both the common spinning lidars (Ouster/Velodyne) and prism-based box-shaped lidar (Livox). {We find that direct method combined with the filter-based update scheme achieves quite good result (\cite{xu2021fast, xu2022fast, lin2021r2live, lin2022r3live}) for a smaller computation footprint compared to more computationally demanding iterative gradient-based optimization.}

In another approach, high-abstraction features such as planes, lines \cite{wisth2021unified, wisth2022vilens} can be extracted and treated as landmarks. {The landmarks' coefficients will also be jointly estimated with the robot states, in similar fashion with visual-inertial odometry (VIO) systems}. This approach has the benefit of dealing with a smaller amount of features. However, the feature extraction process is more complex and may not be able to detect planes or lines in cluttered or unstructured environments such as forests or mines.

\begin{figure}
    \centering
    \begin{overpic}[width=0.95\linewidth,
                        ]{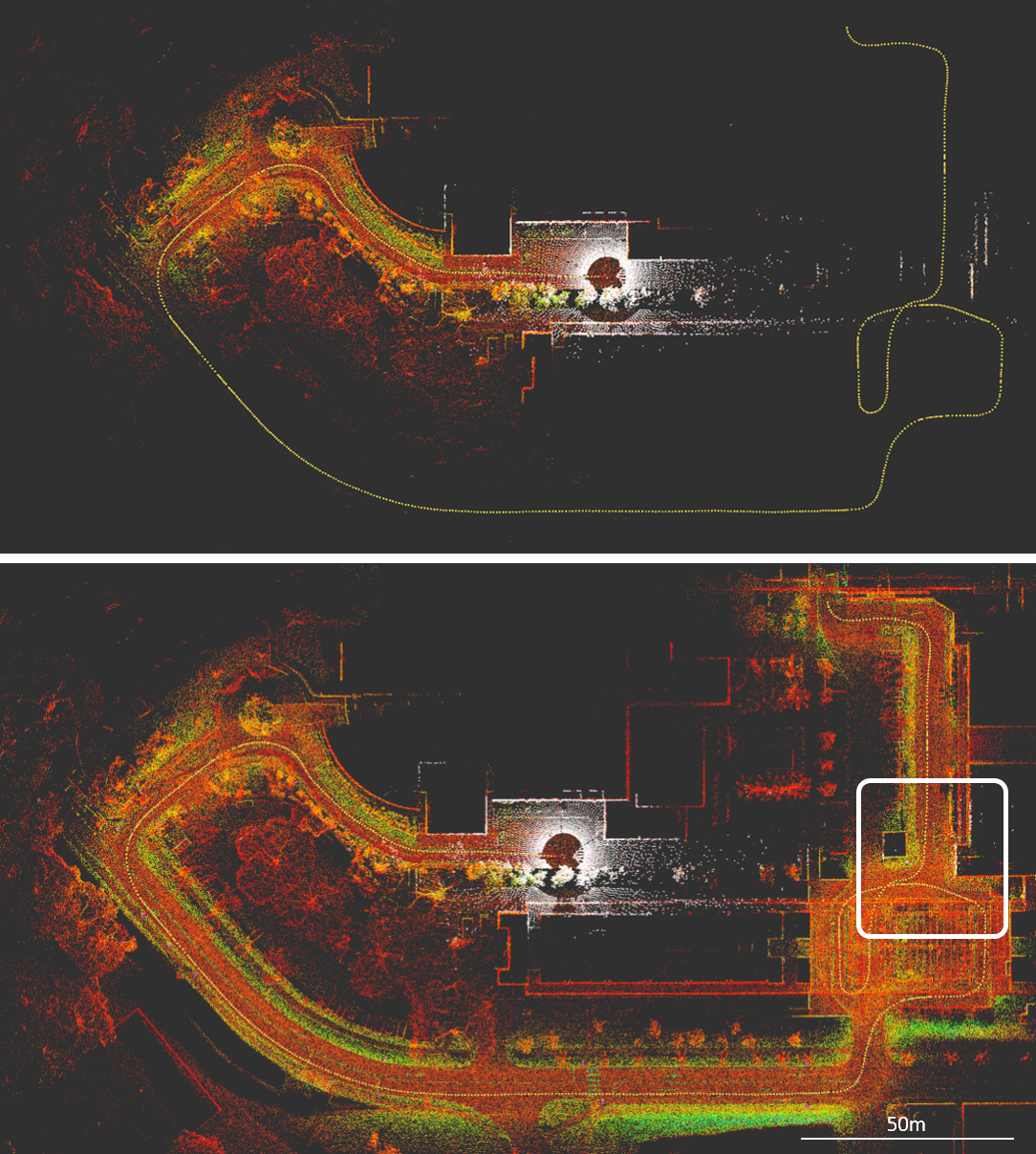}
	\end{overpic}
	\caption{While a local map (top) helps keep computation bounded, it limits the association only to recent marginalized features, which gives chance to estimation drift relative to the initial state. In contrast, with a global map, we can achieve early association with the earliest features (white box) and reinforce global consistency.}
	\label{fig: local vs global}
\end{figure}

In this work, we propose the use of surfel-based features at different resolutions for the frontend task. Different from geometric features such as planes or edges, surfels are characterized by ellipsoidal fitting parameters \cite{quenzel2021real}, with complexity between that of the direct method and high-abstraction features. Moreover, surfels can be detected and associated at different scales, which allow us to add more factors in the graph optimization to reinforce the registration of lidar scans with large scale surfaces in the map.
Instead of associating surfel to surfel like previous works \cite{quenzel2021real, bosse2012zebedee, bosse2009continuous}, which requires complicated processes that involve surfelizing the input scans, then associating up-to-scale surfels using k-NN search based on high dimensional descriptors, and rejecting some associations based on timestamps and velocities \cite{bosse2012zebedee}, we propose a simpler strategy that directly matches the lidar points to surfels at all possible scales, as long as a high fitting score is met (Sec. \ref{sec: association}).

On the mapping side, the update and query processes require special attention. In most conventional systems, a k-d tree structure is built to accelerate the k-NN querying process. However, as the map grows, recomputing the k-d tree on the global map is intractable. One strategy to avoid this issue is to organize the maps into keyframes and build a local map with bounded size from a finite number of nearest keyframes \cite{shan2020liosam, nguyen2021miliom, chen2021low, lv2021clins, quenzel2021real}. However this strategy is sub-optimal since there are cases where the local map misses important prior observations that were captured in previous keyframes (an example is given in Fig. \ref{fig: local vs global}).
To maintain efficient queries on a single global map, in \cite{xu2022fast}, Xu et al proposes the ikd-Tree framework, which allows one to incrementally update the k-d tree of the map without having to recompute all distances from scratch. So far, the ikd-Tree only manages point-based incrementally updated map.
In this paper, we propose a more generalized incremental tree of surfels, based on the UFOMap framework \cite{duberg2020ufomap}.

The contributions in our work can be listed as follows:

\begin{itemize}
    \item A full-fledged Lidar-Inertial Odometry and Mapping framework with frontend odometry, loop closure and global pose-graph optimization, capable of integrating multiple lidar inputs.
    \item A multi-scale point-to-surfel (PTS) association strategy and contiuous-time optimization of PTS and IMU preintegration factors on a sliding window.
    \item An implementation of a hierarchical multi-scale surfel map based on the UFOMap framework, enabling incremental update and efficient query of the global map.
    \item Extensive experiments on public and in-house datasets to validate the results and effects of the multi-scale surfel association strategy.
    \item We release the source code and the datasets for the benefit of the community.
\end{itemize}

Henceforth, we refer to our method as SLICT (\textit{\ul{S}urfel-based \ul{L}idar-\ul{I}nertial \ul{C}ontinous-\ul{T}ime Odometry and Mapping}) for short. The remaining of the paper is as follows: Sec. \ref{sec: prelim} provides some preliminary definitions for our problems, where Sec. \ref{sec: surfel and UFO map} details the surfel tree and its implementation on the UFOMap. Sec. \ref{sec: methodoldy} describes in details the operation of the main blocks in our system. Sec. \ref{sec: exp} describes our experiments on public and in-house datasets. Sec. \ref{sec: conclusion} concludes our work and provides comments for future extension.

\section{Preliminary} \label{sec: prelim}

\subsection{Notation}

In this paper for a vector $\pos \in \R^3$, $\pos^\top$ denotes its transpose, and $\sksym{\pos}$ denotes the skew-symmetric matrix of $\pos$.
For a physical quantity $\pos$, we use the hat notation $\hat{\pos}$ to denote the optimization-based estimate of $\pos$. Similarly, the breve notation $\breve{\pos}$ is used to refer to an IMU-propogated estimate of $\pos$.
To avoid convoluting definitions, we may refer to $\hat{\pos}$, $\breve{\pos}$ without first formally defining $\pos$.

The robot orientation can be represented by the rotation matrix, denoted as $\rot$, or quaternion $\quat$. The two representations can be used interchangeably as the conversion between them is well defined.

We reserve the left superscript for the coordinate frames of a quantity. For example, ${}^{B_{t_s}}\f$ denotes a vector $\f$ whose coordinates are referenced in the robot's body at time $t_s$. For convenience we may also write ${}^{B_{t_s}}\f$ as ${}^{t_s}\f$.

\subsection{State estimate}

We define a sliding window spanning a time interval $[t_w, t_k]$, with $M$ time instances ($t_w$, $t_k$ included). Each $t_m$ is associated with a state estimate $\hat{\X}_m$ defined as follows:
\begin{equation}
    \hat{\X}_m = (\hat{\rot}_m, \hat{\pos}_m, \hat{\vel}_m, \hat{\bias}_{gm}, \hat{\bias}_{am}) \in \SO \times \R^{12},
\end{equation}
where $\hat{\rot}_m \in \SO, \hat{\pos}_m, \hat{\vel}_m \in \R^3$ are respectively the state estimates of the rotation matrix, position and velocity of the robot, and $\hat{\bias}_{gm} \in \R^3, \hat{\bias}_{am} \in \R^3$ are the IMU gyroscope and accelerometer biases.

\subsection{Maximum-A-Priori (MAP) Optimization}

The core of our system lies in solving the following MAP optimization problem:
\begin{align}
    &f(\hat{\X}) = \sum_{m = w}^{k-1}\norm{r_\I(\I_m, \hat{\X}_m, \hat{\X}_{m+1})}^2_{\S_\I} \nonumber\\
    &+ \sum_{m = w}^{k-1}\sum_{\L \in \mathcal{A}_m}\norm{r_\L(\L({}^{B_{t_s}}\f, \n, \mu, s), \hat{\X}_m, \hat{\X}_{m+1})}^2_{\S_\L}
    , \label{eq: cost function}
\end{align}
where:
\begin{itemize}
    \item $r_\I$ is the cost factor of the preintegration $\I_{m}$, which is constructed from the IMU samples in the interval $[t_m, t_{m+1}]$, and $\S_\I$ is the corresponding covariance matrix;
    \item $r_\L$ denotes a lidar PTS factor based on a tuple of PTS association coefficients $\L({}^{B_{t_s}}\f, \n, \mu, s)$, and $\S_\L$, a covariance, which is scalar in this case;
    \item $\n$, $\mu$ are the normal and mean of the underlying associated surfel (Sec. \ref{sec: surfel attributes}), and $ s = \frac{t_s - t_m}{t_{m+1} - t_m}$ is the normalized time stamp of raw lidar point ${}^{B_{t_s}}\f$ (elaborated in \ref{sec: association}).
    \item Finally $\mathcal{A}_m$ denotes the set of successful PTS associations in $[t_m, t_{m+1}]$ {(more details are given in \ref{sec: association})}.
\end{itemize}

It should be noted that each point ${}^{B_{t_s}}\f$ can be associated with multiple surfels at different scales, as long as the surfel satisfies the association predicates in Sec. \ref{sec: association}.

{We refer to our method as continuous-time LIOM based on the use of continuous-time factor $r_\L(\L({}^{B_{t_s}}\f), \hat{\X}_m, \hat{\X}_{m+1})$, which is based on raw lidar point ${}^{B_{t_s}}\f$. This characterization is consistent with previous works \cite{lv2021clins, quenzel2021real}}. In contrast, discrete-time methods use the deskewed point ${}^{B_{t_m}}\f$ based on IMU-propagated states \cite{ye2019tightly, shan2020liosam, nguyen2021miliom}. To couple ${}^{B_{t_s}}\f$ with the states $\hat{\X}_m, \hat{\X}_{m+1}$, we use a linear interpolation formulation based on the time stamp $t_s$ of ${}^{B_{t_s}}\f$, which is elaborated in Sec. \ref{sec: association}.

\subsection{Surfel and UFOMap} \label{sec: surfel and UFO map}

\begin{figure}
    \centering
    \begin{overpic}[width=0.75\linewidth,
                        ]{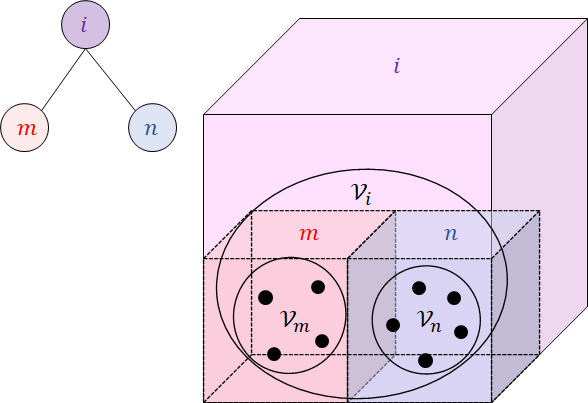}
	\end{overpic}
	\caption{ Illustration of surfel and the tree-like structure: A voxel $i$ contains a set of points $\mathcal{V}_i$, which (in this example) contains two subsets $\mathcal{V}_m$, $\mathcal{V}_n$, encapsulating the points in the voxels $m$, $n$ at smaller scales. Each of the sets $\mathcal{V}_i$, $\mathcal{V}_m$, $\mathcal{V}_n$ define a surfel whose attributes are described in Section \ref{sec: surfel attributes}. As $\mathcal{V}_m \subset \mathcal{V}_i$, $\mathcal{V}_n \subset \mathcal{V}_i$, there is a hierarchical relationship between the surfels, which we propose to organize in an octree implemented by the UFOMap framework.}
	\label{fig: surfel illustration}
\end{figure}

\subsubsection{Surfel's attributes} \label{sec: surfel attributes}

A node $i$ in the octree-based UFOMap corresponds to a voxel, and the node's depth indicates the voxel's scale. For leaf nodes, they are assigned depth 0, its parent has depth 1, and so forth.
A surfel at node $i$ in the UFOMap is defined by the set of points contained within the voxel, denoted as $\V_i = \{\f_1, \dots \f_{N_i}\}$. The following quantities make up the basic attributes of the surfel:
\begin{equation}
    N_i = |\V_i|,\ 
    \Smean_i \triangleq \sum_{k=1}^{N_i} \f_k,\
    \Scov_i  \triangleq \sum_{k=1}^{N_i} \f_k\f_k^\top - \frac{1}{N_i}\Smean_i\Smean_i^\top.
\end{equation}

Each surfel in UFOMap stores the values $N_i$, $\Smean_i$, $\Scov_i$. Given these attributes, we can quickly compute the mean $\mu_i$ and covariance $\G_i$  of $\mathcal{V}_i$, along with other quantities as follows:
\begin{align}
    &\mu_i \triangleq \frac{1}{N_i} \Smean_i,\ 
    \G_i  \triangleq \frac{1}{N_i - 1}\Scov_i,\ 
    \lambda_0 \leq \lambda_1 \leq \lambda_2, \nonumber\\
    &\rho_i = 2 \frac{\lambda_1 - \lambda_0}{\lambda_0 + \lambda_1 + \lambda_2},\ \n_i \triangleq \nu_0 \nonumber,\ c_i \triangleq -\nu_0^\top\mu_i.
\end{align}
where $\mu_i$ is the mean, $\G_i$ is the covariance with the eigenvalues $\lambda_0$, $\lambda_1$, $\lambda_2$ and $\n_i = \nu_0$ is the normalized eigenvector associated with $\lambda_0$; $\rho_i$ is the so-called \textit{planarity} value, i.e. the plane-likeness metric of the surfel.

Besides the aforementioned attributes, the \textit{depth} and \textit{scale} of the voxel node are also frequently queried. To initialize a surfel map in the UFOMap framework, we assign a \textit{leaf node size}, denoted $\ell$, for the voxels at the smallest scale, i.e., at depth 0. The scale of a voxel at depth $D$ is therefore $2^D\ell$. 

\subsubsection{Surfel addition and subtraction} \label{sec: surfel add and sub}

Assuming that a parent node $i$ has two children $m$ and $n$, the surfel attributes of $i$ can be calculated via Welford's formula \cite{welford1962note}:
\begin{align}
    &\a         \coloneqq 1/\l[N_{m}N_{n}(N_{m} + N_{n})\r],\ \b \coloneqq N_n\Smean_{m} - N_m\Smean_{n}, \nonumber \\
    &\Scov_{i}  \doteq \Scov_{m} + \Scov_{n} + \a\b\b^\top, \nonumber\\
    &N_i        \doteq N_m + N_n,\ \Smean_{i} \doteq \Smean_{m} + \Smean_{n}, \label{eq: welford add}
\end{align}
The above can be iterated for parent nodes with more than two children.

Based on \eqref{eq: welford add}, if the child node $n$ is removed from the map, the surfel attributes of $i$ can be updated by the procedure:
\begin{align}
    &N_i       \doteq N_i - N_n,\ \Smean_{i} \doteq \Smean_{i} - \Smean_{n}, \nonumber\\
    &\a        \coloneqq 1/\l[N_{i}N_{n}(N_{i} + N_{n})\r],\ \b \coloneqq N_n\Smean_{i} - N_i\Smean_{n}, \nonumber\\
    &\Scov_{i} \doteq \Scov_{i} - \Scov_{n} - \a\b\b^\top. \label{eq: welford subtract}
\end{align}

For single point update, i.e. the update at the leaf nodes when new pointcloud is added to the UFOMap, we can still use \eqref{eq: welford add}, \eqref{eq: welford subtract} by noticing that when $N_n = 1$, $\Scov_n = 0$.

\section{System Description} \label{sec: methodoldy}

\begin{figure}
    \centering
    \begin{overpic}[width=\linewidth,
                        ]{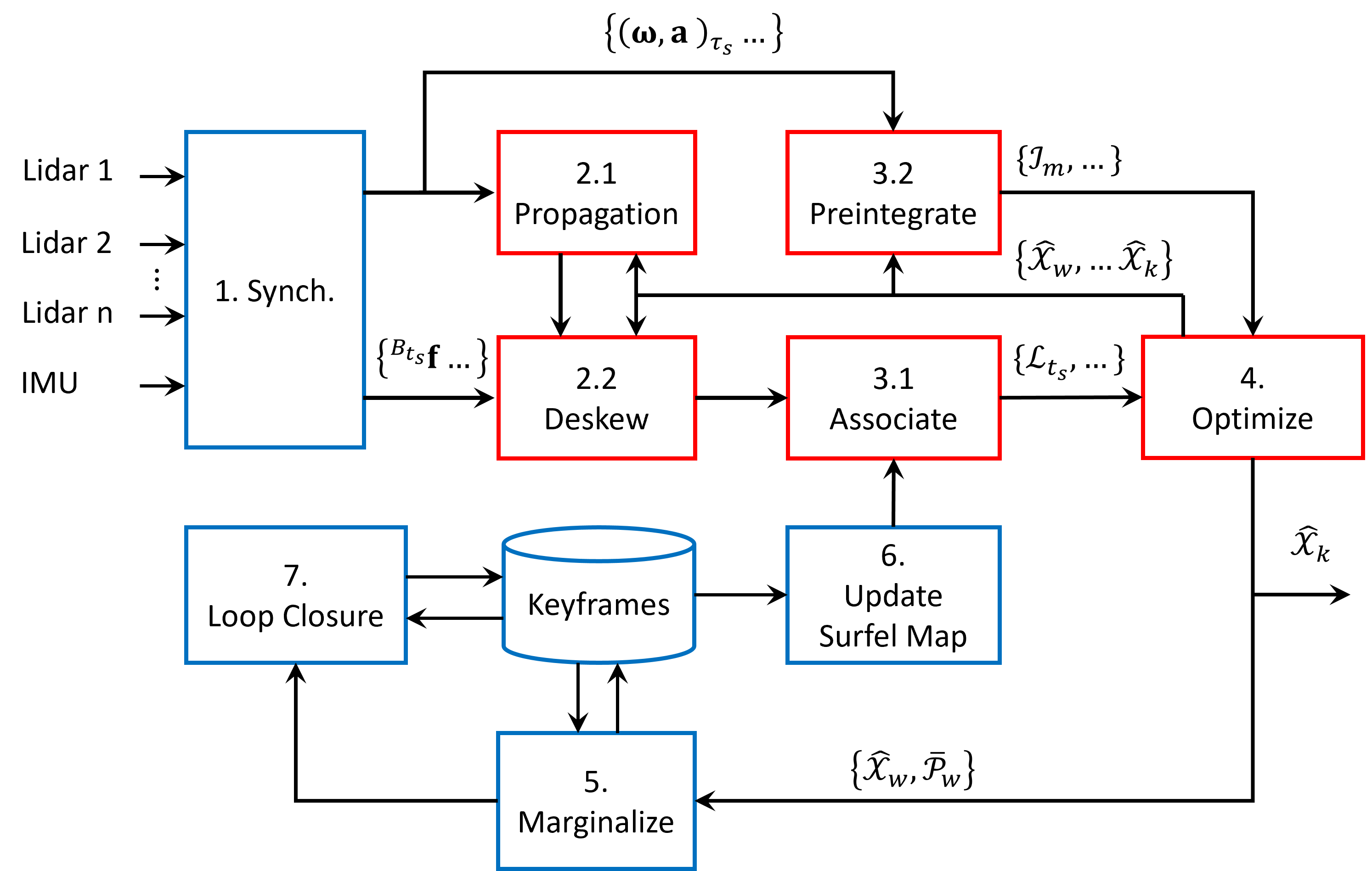}
	\end{overpic}
	\caption{The general workflow of the system.
	More details are given in Sec. \ref{sec: sync} - Sec. \ref{sec: loop closure}}
	\label{fig: workflow}
\end{figure}

Fig. \ref{fig: workflow} provides an overview of our system. In the next subsections we describe in details each numbered block.

\subsection{Synchronization} \label{sec: sync}

Synchronization is a prerequisite to optimization based estimation methods. Fig. \ref{fig: sync} gives a brief explanation of our synchronization scheme. More details are given below.

The sliding window's progression is based on the arrival time of new lidar messages from a so-called \textit{primary} lidar. The messages from each lidar are stored in an ordered buffer such that data from each lidar can be treated as a continuous stream. As the window shifts from $[t_w - \D_l, t_k - \D_l]$ to $[t_w, t_k]$, (where $\D_l$ is the sweeping period of the primary lidar, typically $0.1s$), we extract data points from all the lidar streams in the period $[t_k - \D_l, t_k]$ and merge them into a unified pointcloud $\P_k$
.

Similarly, IMU samples are also stored in a buffer, and when the window slides forward for $\D_l$ seconds, the IMU samples in the periods $[t_k - \D_l, t_k]$ are also extracted and put onto the sliding window.

When a new lidar pointcloud is admitted to the sliding window, we also add one or more state estimates to the sliding window.
Thus the number of pointclouds can be smaller than the number of state estimates, which is different from our previous work MILIOM \cite{nguyen2021miliom}. The addition of more than one state estimates per newly added pointcloud is to better capture the dynamics of the robot.
In practice we add 2 to 8 new state estimates to the sliding window, depending on the frequency of the IMU.

\begin{figure}
    \centering
    \begin{overpic}[width=\linewidth,
                        ]{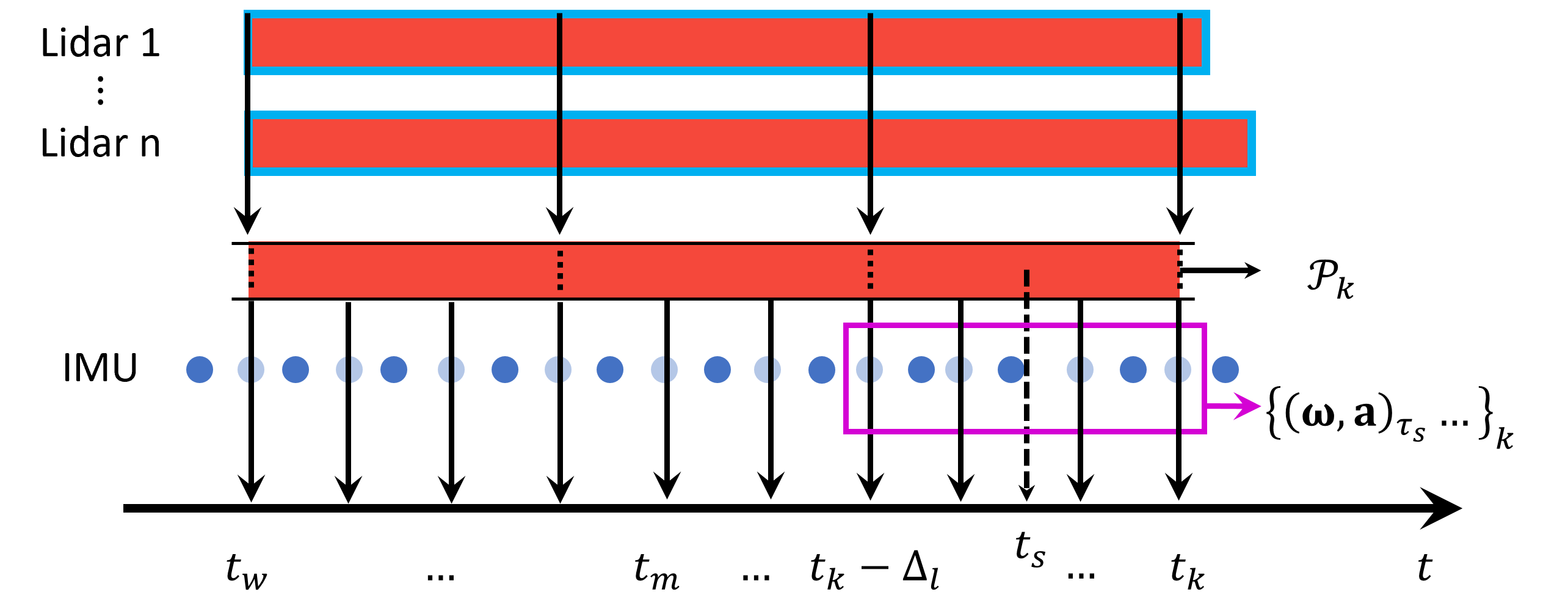}
	\end{overpic}
	\caption{ Synchronization scheme on the sliding window: the window is shifted forward by $\D_l$ seconds based on the arrival time of the primary lidar messages. Data from each lidar is treated as a continuous stream of points. At time $t_k$, we extract and merge all points that fall in the period $[t_{k} - \D_l, t_k)$ into a single pointcloud $\mathcal{P}_k$. Similarly, IMU samples during $[t_{k} - \D_l, t_k)$ are also extracted and associated with $\mathcal{P}_k$. On the sliding window, we define a number of time instances $t_m \in [t_w, t_k]$, each is associated with a state estimate $\hat{\X}_m$. For each lidar point, we use their time stamp $t_s$ to find the interval $[t_m, t_{m+1}) \ni t_s$, and the same for IMU sample. The interval $[t_m, t_{m+1})$ determines the state estimates $\hat{\X}_m, \hat{\X}_{m+1}$ that are coupled with the lidar or IMU measurements in \eqref{eq: cost function}.}
	\label{fig: sync}
\end{figure}

\subsection{IMU propagation} \label{sec: imu prop}

Given the IMU samples $\{(\angvel, \accel)_{\tau_i}, \dots\}$, $\tau_i \in [t_m, t_{m+1}]$, and the starting state $\hat{\X}_m := \breve{\X}_{m}$, we can forward propagate the robot state to $\breve{\X}_{m+1}$, or backward propagate from $\hat{\X}_{m+1}$ to $\breve{\X}_m$. After this process we have a sequence of IMU-propagated states $\{\breve{\X}_{\tau_i}\dots\}$, $\tau_i \in [t_m, t_{m+1}]$. This sequence can be used to initialize the new state estimate added to the sliding window, or to remove the motion-induced distortion of the pointcloud. We explain this so-called \textit{deskew} process in the next part.

\subsection{Deskew (motion compensation)}

For each lidar point ${}^{B_{t_s}}\f$, $t_s \in [t_m, t_{m+1}]$ we seek to transform its coordinate to the body frame at time $t_{m+1}$. To this end, we search for the two IMU-propagated poses closest to $t_s$, i.e. $\breve{\tf}_{\tau_a}, \breve{\tf}_{\tau_b}$, where $\tau_a \leq t_s \leq \tau_b$, and find the linearly interpolated pose $\breve{\tf}_{t_s}$:
\begin{equation}
    \breve{\tf}_{t_s}=\begin{bmatrix}\mathrm{slerp}(\breve{\rot}_{\tau_a}, \breve{\rot}_{\tau_b}, s) &(1 - s)\pos_{\tau_a} + s\pos_{\tau_b}\\ 0 &1\end{bmatrix},
\end{equation}
where $s \triangleq \frac{t_s - \tau_a}{\tau_b - \tau_a}$ and $\mathrm{slerp}()$ denotes the spherical linear interpolation operation on $\SO$.

Given $\breve{\tf}_{t_s}$, we can transform ${}^{B_{t_s}}\f$ to the world frame by $\f = \breve{\rot}_{t_s}{}^{B_{t_s}}\f + \hat{\pos}_ts$. Hence we proceed to associating these lidar points with the surfel map.

\subsection{Point-to-Surfel Association} \label{sec: association}

The association consists of two stages.
In the first stage, for each lidar point $\f$ we find all the nodes $\V_i$ matching the following predicates in the surfel map:
\begin{itemize}
    \item The voxel depth is between 1 and $D_\mathrm{max}$ (the leaf nodes are ignored).
    \item $N_i \geq N_\text{min}$ and $\rho_i > \rho_\text{min}$, i.e. $\V_i$ should have at least $N_\text{min}$ points and the planarity is sufficiently large.
    \item The voxel's cube intersects with a sphere of radius $r > 0$ centered at $\f$, { where $r$ is a user-defined parameter}.
\end{itemize}

If a surfel $\V_i$ passes all of the aforementioned predicates, in the second stage we calculate the distance of $\f$ to the underlying plane of $\V_i$, i.e. $d_i = \n_i^\top(\f-\mu_i)$. If $d_i < d_\text{max}$, a tuple of PTS coefficients $\L = ({}^{t_s}\f, \n_i, \mu_i, s),\ s = \frac{t_s - t_m}{t_{m+1} - t_m}$ will be added to the set of successful association $\mathcal{A}_m$, which is then used to construct the cost function at the optimization stage \eqref{eq: cost function}.

Given $\L$, the cost factor $r_\L$ in \eqref{eq: cost function} can be constructed as follows:
\begin{align}
    r_\L &= \n^\top(\hat{\rot}_s{}^{t_s}\f + \hat{\pos}_s - \mu), \nonumber\\ 
    \hat{\rot}_s &= \hat{\rot}_m\Exp\l[s\Log(\hat{\rot}_m^{-1}\hat{\rot}_{m+1})\r], \nonumber\\  
    \hat{\pos}_s &= (1 - s)\hat{\pos}_{m} + s\hat{\pos}_{m+1}.
\end{align}

\subsection{IMU Preintegration}

We refer to our previous work \cite{nguyen2021miliom, nguyen2021viral} for the detailed formulation and Jacobian of IMU-preintegration factors in optimization-based estimation.

\subsection{Optimization} \label{sec: optimization}

Once all of the PTS coefficients and IMU preintegrations have been prepared, we proceed to construct and optimize the cost function \eqref{eq: cost function} using the ceres solver \cite{ceres-solver}.
It should be noted that the steps described in Sec. \ref{sec: imu prop} to Sec. \ref{sec: optimization} (red boxes in Fig. \ref{fig: workflow}) can be done iteratively.

\subsection{Marginalization}

After each optimization step, we check if the earliest pose estimate $\hat{\tf}_w$ can be marginalized to become a keyframe. To this end, we search for five nearest neighbours of $\hat{\tf}_w$ among the existing keyframes. If the Euclidean distance or rotational distance of $\hat{\tf}_w$ to all of its neighbours exceeds a threshold, its associated deskewed pointcloud will be inserted to the surfel map. We also marginalize $\hat{\tf}_w$ and store its deskewed pointcloud in the buffer to be reused in case of loop closure.

\subsection{Updating the surfel map}

When a new keyframe is admitted to the buffer, we also update the surfel map using the procedure that was introduced in Sec. \ref{sec: surfel add and sub}.

\subsection{Loop Closure} \label{sec: loop closure}

When a new keyframe with the pose estimate $\hat{\tf}_c$ is admitted to the buffer, we will find $K$ nearest keyframes of $\hat{\tf}_c$. If the time stamp of one keyframe $\hat{\tf}_p$ and $\hat{\tf}_m$ differs by a certain amount, it signals the return to a previously explored area. We use the ICP algorithm to calculate the relative pose ${}^{p}\bar{\tf}_c$ between the two keyframes and store this in a buffer $\T$.

When a new loop is admit, we optimize the pose graph with this loop prior:
\begin{equation}
    = \sum_{k = 0}^{K-1} {}^{k}\bar{\tf}_{k+1} \hat{\tf}_k^{-1}\hat{\tf}_{k+1} + \sum_{{}^{p}\bar{\tf}_c \in \mathcal{T}} {}^{p}\bar{\tf}_c\hat{\tf}_p^{-1}\hat{\tf}_{c},
\end{equation}
where ${}^{k}\bar{\tf}_{k+1}$, ${}^{p}\bar{\tf}_{c}$ are the relative pose priors, and $\mathcal{T}$ is the set of relative poses detected. After the pose graph optimization, the global map will be recomputed.

\section{Experiments} \label{sec: exp}

We demonstrate the performance of SLICT via three data suites: NTU VIRAL \cite{nguyen2021ntuviral}, Newer College \cite{ramezani2020newer} and our in-house datasets, which respectively capture operations at different scales.

We compare our methods with three other state-of-the-art methods. First is MARS \cite{quenzel2021real}, which stands for Multi-Adaptive-Resolution-Surfel with B-spline-based continuous-time optimization. Second is LIO-SAM \cite{shan2020liosam}, which uses conventional plane-edge feature and IMU preintegration factors, optimized by the Georgia Tech Smoothing and Mapping framework \cite{gtsam}. Third is FAST-LIO2, which uses the direct method for feature association and the ikd-Tree global mapping scheme.

\subsection{NTU VIRAL Datasets}

The NTU VIRAL is a multi-lidar dataset collected from an Unmanned Aerial Vehicles, with ground truth of centimeter-level accuracy obtained from a laser-tracker total station. The environments consist of indoor and outdoor spaces where the UAV operates within a volume of 50m radius.

For all experiments, we merge the data from the so-called horizontal and vertical lidars in the NTU VIRAL dataset and use them as the input of all four methods. All experiments are run on a core-i7 PC. The metric used in this case is the Absolute Trajectory Error (ATE).

\begin{table}
\centering
\caption{ATE of SLICT compared with other methods on NTU VIRAL datasets (unit [m]). The best results are in \tb{bold}, second best are \ul{underlined}. 'x' denotes divergence.}
\label{tab: ATE NTU VIRAL}
\begin{tabular}{cccccc}
\hline\hline
\tb{Dataset}
&\tb{MARS}
&\tb{\begin{tabular}[c]{@{}c@{}}LIO-\\SAM\end{tabular}}
&\tb{\begin{tabular}[c]{@{}c@{}}FAST-\\LIO2\end{tabular}}
&\tb{SLICT}
\\ \hline
{eee\_01}
        &{0.2471}			
        &{0.0624}			
        &\ul{0.0585}			
        &\tb{0.0316}\\		
{eee\_02}
        &{0.1033}			
        &{0.0457}			
        &\ul{0.0318}			
        &\tb{0.0249}\\		
{eee\_03}
        &{0.0927}			
        &{0.0403}			
        &\ul{0.0351}			
        &\tb{0.0275}\\		
{nya\_01}
        &{0.0555}			
        &{2.0960}			
        &\ul{0.0305}			
        &\tb{0.0229}\\		
{nya\_02}
        &{0.0624}			
        &{x}			
        &\ul{0.0286}			
        &\tb{0.0227}\\		
{nya\_03}
        &{0.0831}			
        &{0.0468}			
        &\ul{0.0315}			
        &\tb{0.0260}\\		
{sbs\_01}
        &{0.1370}			
        &{0.0444}			
        &\ul{0.0324}			
        &\tb{0.0298}\\		
{sbs\_02}
        &{0.1256}			
        &{0.0461}			
        &\ul{0.0322}			
        &\tb{0.0291}\\		
{sbs\_03}
        &{0.1588}			
        &{0.0494}			
        &\ul{0.0428}			
        &\tb{0.0335}\\		
{rtp\_01}
        &{x}			
        &{0.2571}			
        &\ul{0.0494}			
        &\tb{0.0447}\\		
{rtp\_02}
        &{0.2329}			
        &\ul{0.1091}			
        &{0.1151}			
        &\tb{0.0466}\\		
{rtp\_03}
        &{0.1377}			
        &{0.0576}			
        &\ul{0.0543}			
        &\tb{0.0501}\\		
{tnp\_01}
        &{0.0734}			
        &{x}			
        &\ul{0.0432}			
        &\tb{0.0287}\\		
{tnp\_02}
        &{0.0681}			
        &\ul{0.0330}			
        &{0.0590}			
        &\tb{0.0201}\\		
{tnp\_03}
        &{0.0665}			
        &\tb{0.0283}			
        &{0.0468}			
        &\ul{0.0383}\\		
{spms\_01}
        &{x}			
        &{0.1620}			
        &\ul{0.0686}			
        &\tb{0.0610}\\		
{spms\_02}
        &{x}			
        &{0.6641}			
        &\tb{0.0821}			
        &\ul{0.1000}\\		
{spms\_03}
        &{19.8650}			
        &{1.0071}			
        &\tb{0.0603}			
        &\ul{0.0661}\\		

\hline\hline
\end{tabular}
\end{table}

In these experiments we do not enable the loop closure function of SLICT and only compare the odometry result.
We set sensor's parameters such as number of lines, minimum distance, IMU noises, etc, according to the dataset's meta data, while other parameters are kept as default. For SLICT, we choose a sliding window of 400ms with 16 intervals, which encompass 4 pointclouds with 4 state estimates each.

\begin{figure}
    \centering
    \begin{overpic}[width=0.9\linewidth,
                        ]{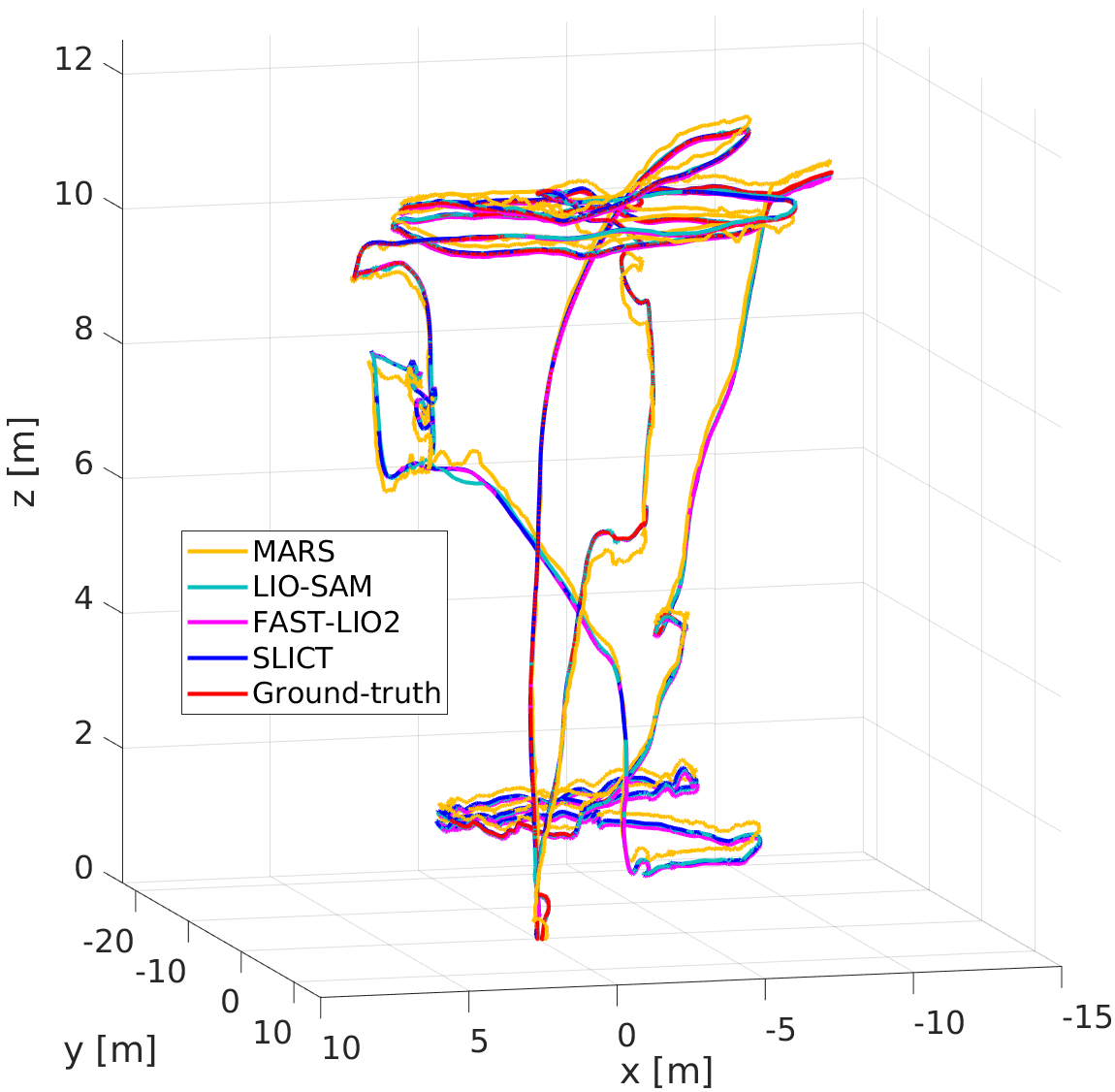}
	\end{overpic}
	\caption{Position estimates by different methods and ground truth in NTU VIRAL sequence rtp\_02. SLICT and FAST-LIO2 with sub-decimeter accuracy follow the ground truth closely, while LIO-SAM and MARS have visible deviation.}
	\label{fig: ntuviral_traj_rtp_02}
\end{figure}

Tab. \ref{tab: ATE NTU VIRAL} reports the result of our experiments. SLICT has the highest accuracy in the most experiments, and FAST-LIO2 has the most second best results. MARS and LIO-SAM diverge in some of the experiments. Fig. \ref{fig: ntuviral_traj_rtp_02} shows the trajectory estimated by different methods and ground truth in sequence rtp\_02.

\begin{figure}
    \centering
    \begin{overpic}[width=0.9\linewidth,
                        ]{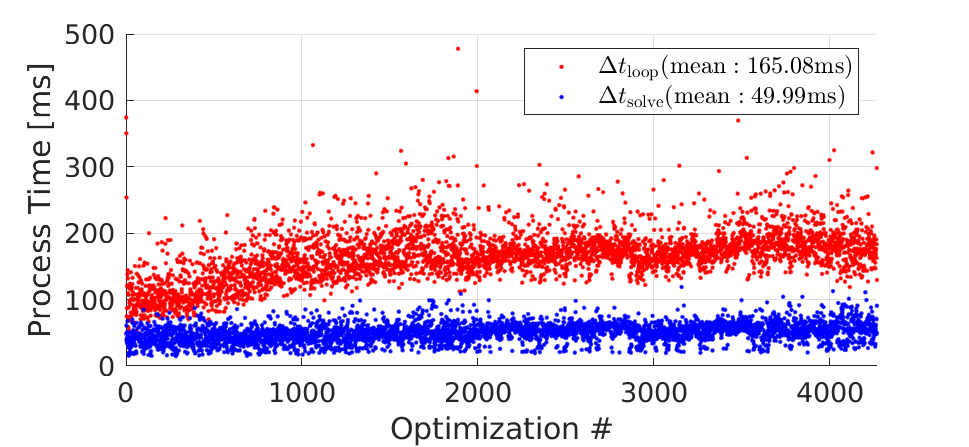}
	\end{overpic}
	\caption{Process time of SLICT in NTU VIRAL sequence nya\_02.}
	\label{fig: ntuviral_nya_02_runtime}
\end{figure}

We analyze the computational load of SLICT in the NTU VIRAL nya\_02 sequence in Fig. \ref{fig: ntuviral_nya_02_runtime}. On average, it takes 165ms to complete one cycle of the algorithm ($\Delta t_{\text{loop}}$, defined as the time from one optimization operatation to another), in which solving the optimization problem takes about 50ms ($\Delta t_{\text{solve}}$), and the rest is for deskew, association, and keyframe marginalization. Since lidar input is acquired at 100ms, currently real-time performance is not guaranteed by SLICT. Because we associate the points with surfels at five scales (from $2^1\ell$ to $2^5\ell$, where $\ell=0.1m$), the computation load for association is at least a multitude that of direct method, which uses only one voxel scale. However, we think it is justifiable considering that SLICT gives higher accuracy, and real-time performance can be achieved by using a CPU that supports more threads.

\subsection{Newer College Dataset}

\begin{figure}
    \centering
    \begin{overpic}[width=0.9\linewidth,
                        ]{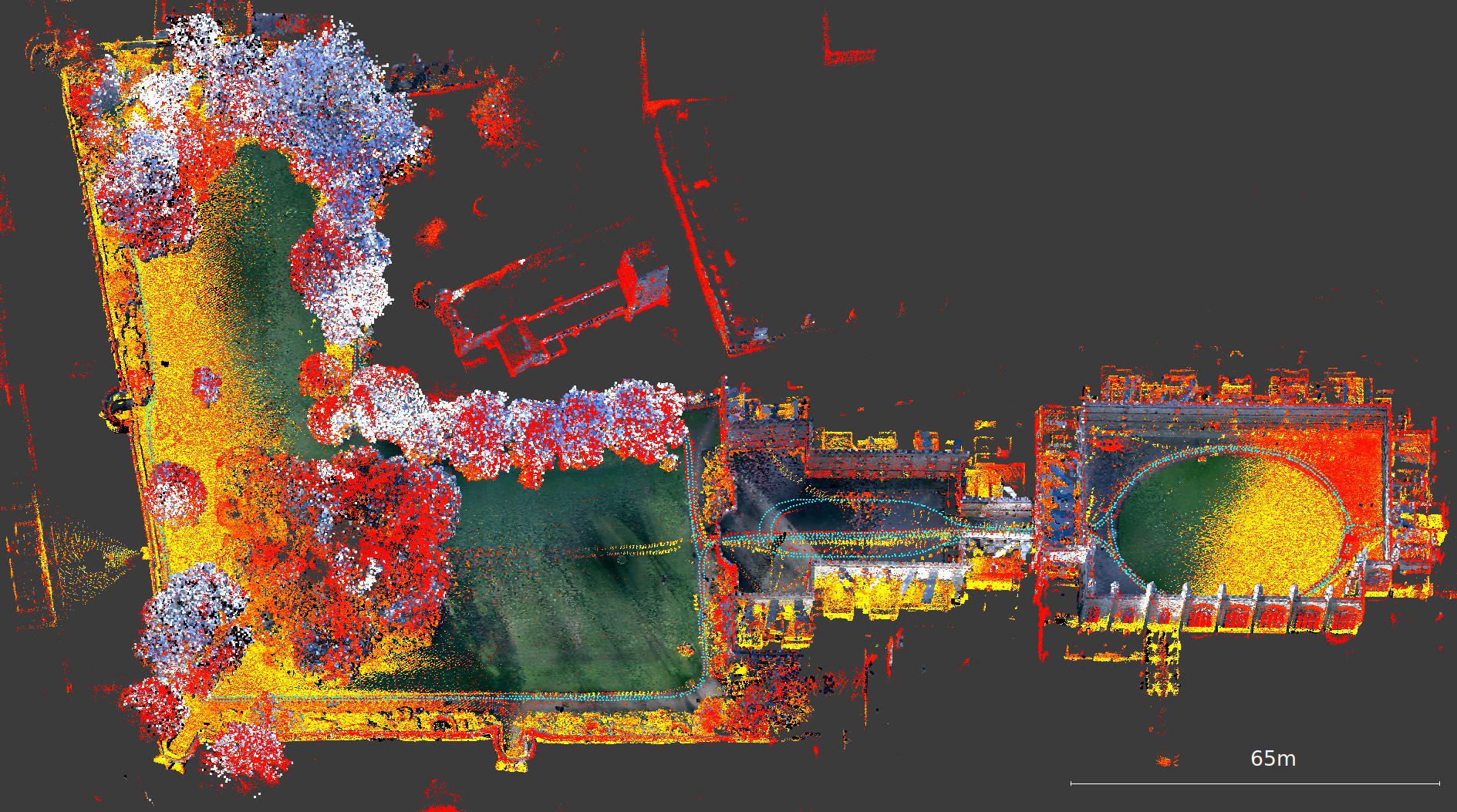}
	\end{overpic}
	\caption{The global map built by SLICT (red-to-yellow gradient based on the points' intensity) and keyframe posistions (cyan dots) from sequence 02 of Newer College Dataset, superimposed on the RGB-colored prior map.}
	\label{fig: oxford global map exp 02}
\end{figure}

The Newer College Dataset consists of five sequences collected at New College, Oxford, featuring an Ouster lidar with 64 channels and 90-degree vertical field-of-view, and a built-in 100Hz IMU. The ground truth is obtained from ICP-based registration of lidar scan with a centimeter-resolution map captured by an imaging lidar scanner. The environment captured in the dataset features two open squares and an open park over an area of roughly 200m $\times$ 100m (Fig. \ref{fig: oxford global map exp 02}), which is several times larger than those in NTU VIRAL sequences, thus localization drift can be observed more easily.

\begin{table}[!h]
\centering
\caption{ATE of SLICT and other methods on Newer College Dataset (unit [m]). The best results are in \tb{bold}, second best are \ul{underlined}. 'x' demotes a divergent experiment.}
\label{tab: ATE Newer College Dataset}
\begin{tabular}{ccccc}
\hline\hline
\tb{Dataset}
&\tb{MARS}
&\tb{\begin{tabular}[c]{@{}c@{}}LIO-\\SAM\end{tabular}}
&\tb{\begin{tabular}[c]{@{}c@{}}FAST-\\LIO2\end{tabular}}
&\tb{SLICT}
\\ \hline
{01\_short\_experiment}
        &{2.1521}			
        &{-}			
        &\ul{0.3883}			
        &\tb{0.3843}\\		
{02\_long\_experiment}
        &{6.0030}			
        &{-}			
        &\ul{0.3659}			
        &\tb{0.3496}\\		
{05\_quad\_with\_dynamics}
        &{0.3729}			
        &{-}			
        &\ul{0.3443}			
        &\tb{0.1155}\\		
{06\_dynamic\_spinning}
        &{x}			
        &{-}			
        &\tb{0.0800}			
        &\ul{0.0844}\\		
{07\_parkland\_mound}
        &{x}			
        &{-}			
        &\ul{0.1356}			
        &\tb{0.1290}\\		

\hline\hline
\end{tabular}
\end{table}

The settings are similar to that in the NTU VIRAL dataset, however for SLICT we assign 2 state estimates for each interval. This is because the dataset's main IMU, which is the built-in IMU of the Ouster lidar, has a frequency of only 100Hz. Note that LIO-SAM does not work in this case as it requires orientation estimate from the IMU.

\begin{figure}
    \centering
    \begin{overpic}[width=0.9\linewidth,
                        ]{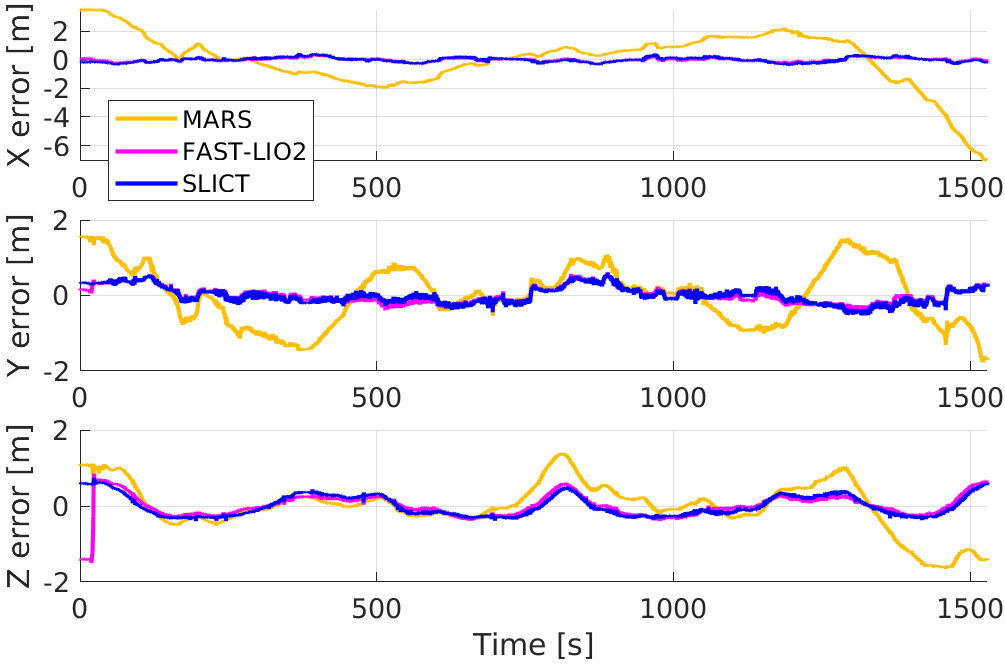}
	\end{overpic}
	\caption{Estimation error over time of the methods in the Newer College Dataset, sequence 01. We find that the error is most significant in the z direction due to the subtle changes in elevation of the area.}
	\label{fig: oxford global map exp 02 err}
\end{figure}

In Tab. \ref{tab: ATE Newer College Dataset}, we report the ATE of the results. Again, it shows that SLICT has the best results in the most experiments, followed closely by FAST-LIO2. MARS's ATE in sequence 01, 02, 05 is similar to the reported result in \cite{quenzel2021real}. As the Newer College sequences 01, 02 and 07 traverse a significantly larger area than the NTU VIRAL dataset, the ATE increases more visibly compared to the NTU VIRAL experiments. Fig. \ref{fig: oxford global map exp 02 err} shows the localization error over time of different methods in sequence 01.

\subsection{In-house Dataset}

\begin{figure*}
    \centering
	\begin{subfigure}[h]{0.32\linewidth}
        \centering
        \includegraphics[width=\linewidth]{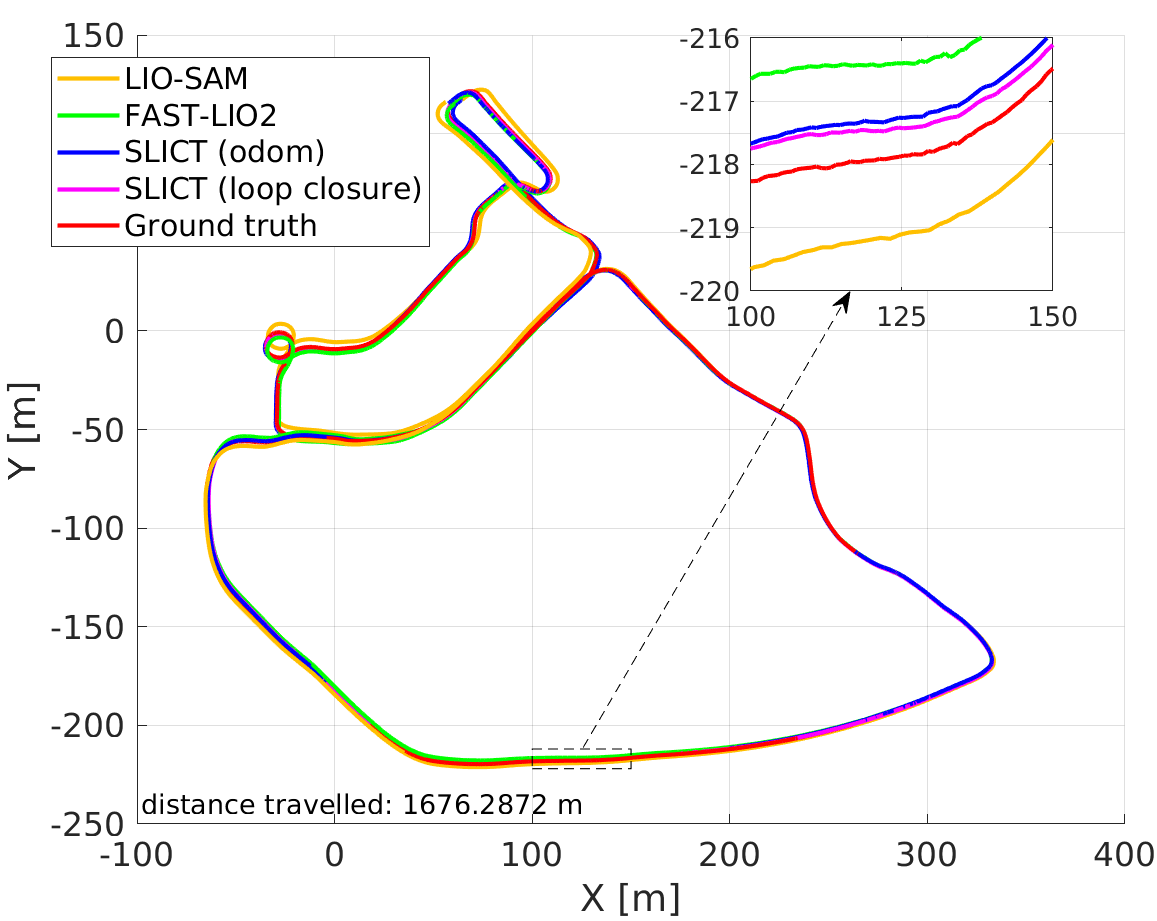}
		\caption{Sequence 1}
		\label{fig: daytime 05}
	\end{subfigure}
	\begin{subfigure}[h]{0.32\linewidth}
        \centering
		\includegraphics[width=\linewidth]{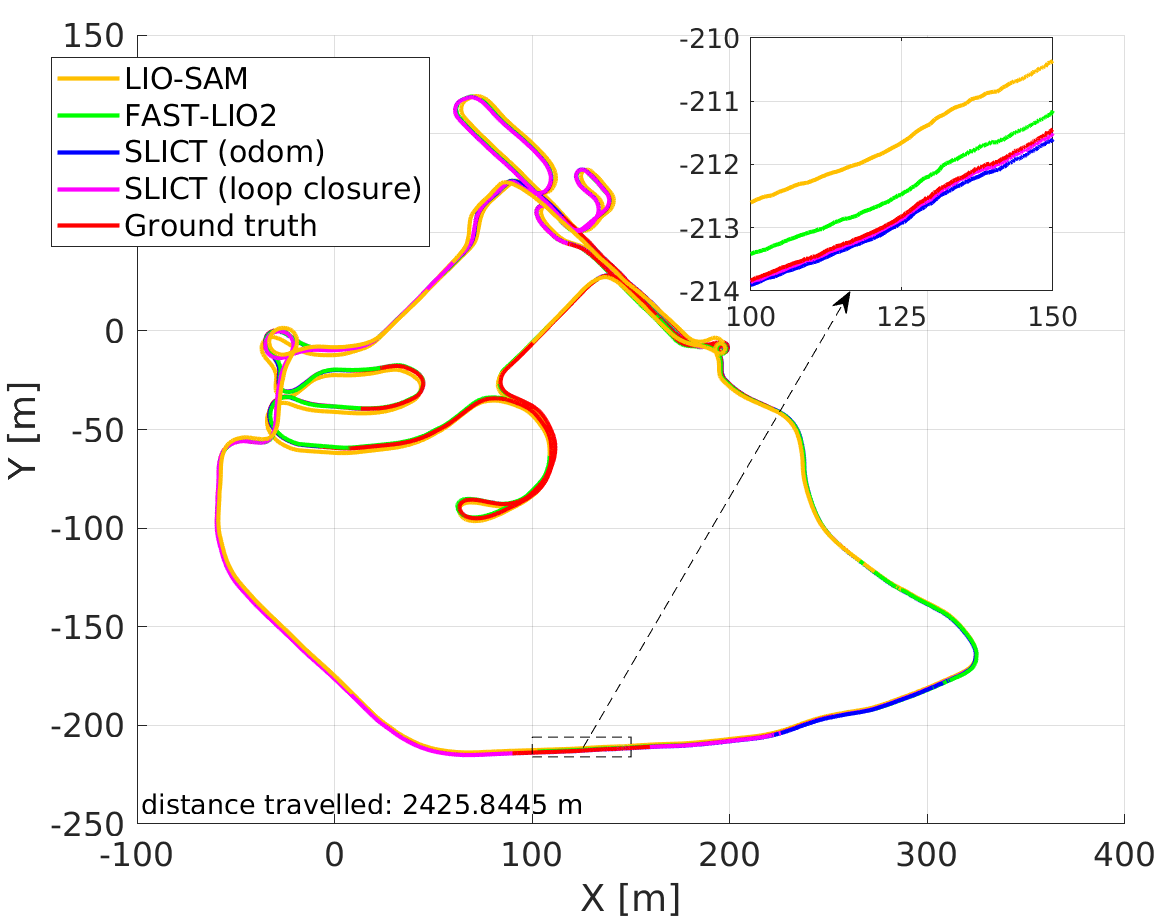}
		\caption{Sequence 2}
		\label{fig: nighttime 08}
	\end{subfigure}
    \begin{subfigure}[h]{0.32\linewidth}
        \centering
        \includegraphics[width=\linewidth]{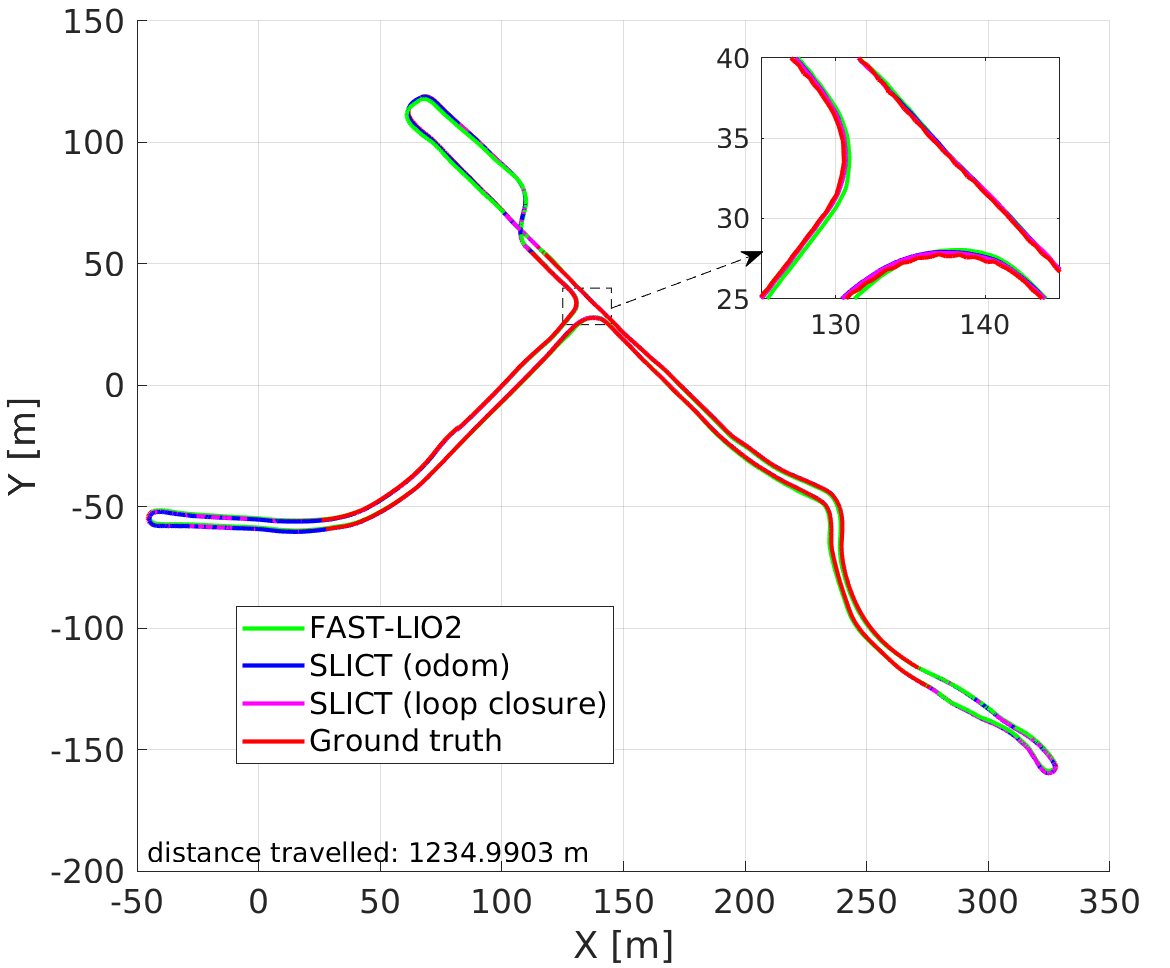}
        \caption{Sequence 3}
		\label{fig: nighttime 13}
	\end{subfigure}
    \caption{Localization estimates from different methods. The zoomed-in plots show the narrow path in the dense-vegetation section (sequences 01 and 02), and the main junction (sequence 03). The elevation difference between these two points are about 13m.}
	\label{fig: mcd traj est}
\end{figure*}

\begin{figure}
    \centering
    \includegraphics[width=0.9\linewidth]{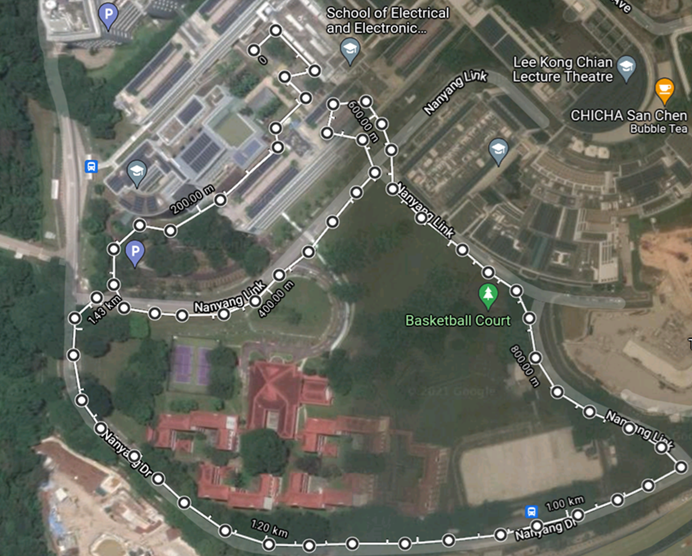}
	\caption{Main routes in the in-house dataset.}
	\label{fig: area of experiment}
\end{figure}

The in-house datasets are collected from a sensor suite consisting of an OS1-128 Lidar, an Livox Mid-70, together with a VectorNav VN100 IMU.
The sensors are mounted on an All-Terrain-Vehicle (ATV) that travels around 1.5 km loop at the southern part of Nanyang Technical University (NTU) Campus (Fig. \ref{fig: area of experiment}). This covers an area of about 400m $\times$ 200m, which is about four times the area of the Newer College dataset. At this scale, the effect of a loop closure module can become more appreciable, thus we add new experiments of SLICT with loop closure and pose-graph optimization functions.

The ground truth of the dataset is constructed in a similar manner to the Newer College Dataset. Specifically, we first build a static map of the environment using the survey scanner Leica RTC360 with centimeter accuracy. We then register the ouster lidar scans with this static map to obtain the ground truth pose at the lidar frequency.

Three sequences are captured. In sequence 1, we traverse the main routes of the environments. In sequence 2, we increase the traversed distance by revisiting some of the routes. In sequence 3, we travel back and forth on the Nanyang Link route to maximize the loop closure incidents. Fig. \ref{fig: mcd traj est} provides an overview of the routes taken.

Similar to previous experiments, we directly merge the lidar sensors into a single input and use it for all methods. Tab. \ref{tab: ATE inhouse dataset} presents the ATE of the tested methods. Due to the high-speed of the ATV (up to 30 km/h), MARS diverges in all three experiments, and LIO SAM diverges in the last experiment. Without loop closure, SLICT still has the best accuracy, and the loop closure improves the accuracy even further. {We also added result of LIO-SAM with loop closure enabled to compare with SLICT}.

\begin{figure}
    \centering
    \begin{overpic}[width=0.9\linewidth,
                        ]{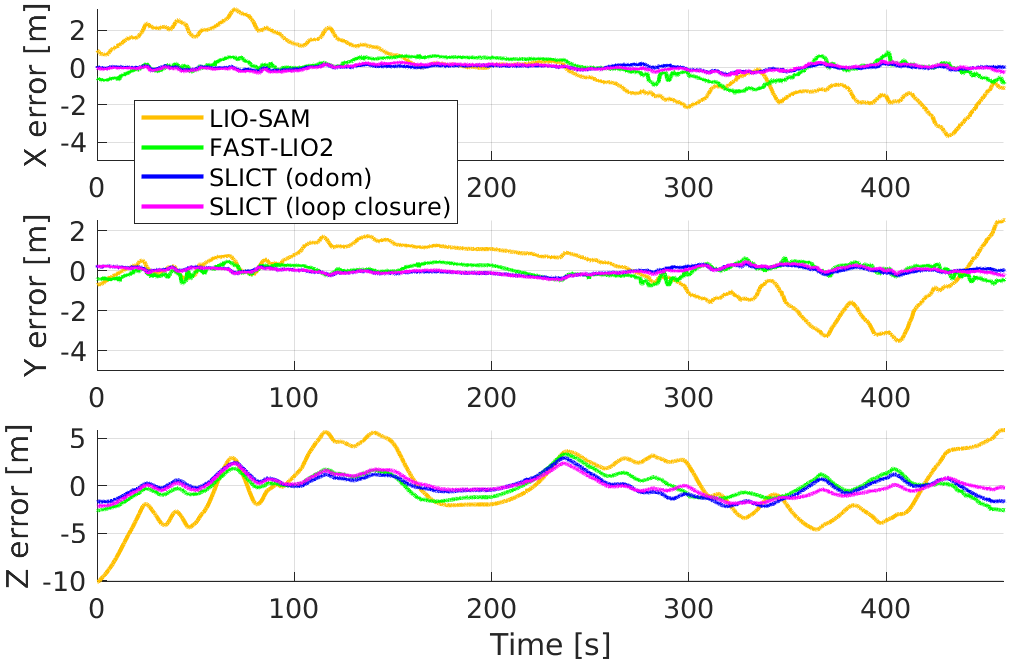}
	\end{overpic}
	\caption{Localization error in sequence 2.}
	\label{fig: mcd error}
\end{figure}

\begin{figure}
    \centering
    \begin{overpic}[width=\linewidth,
                        ]{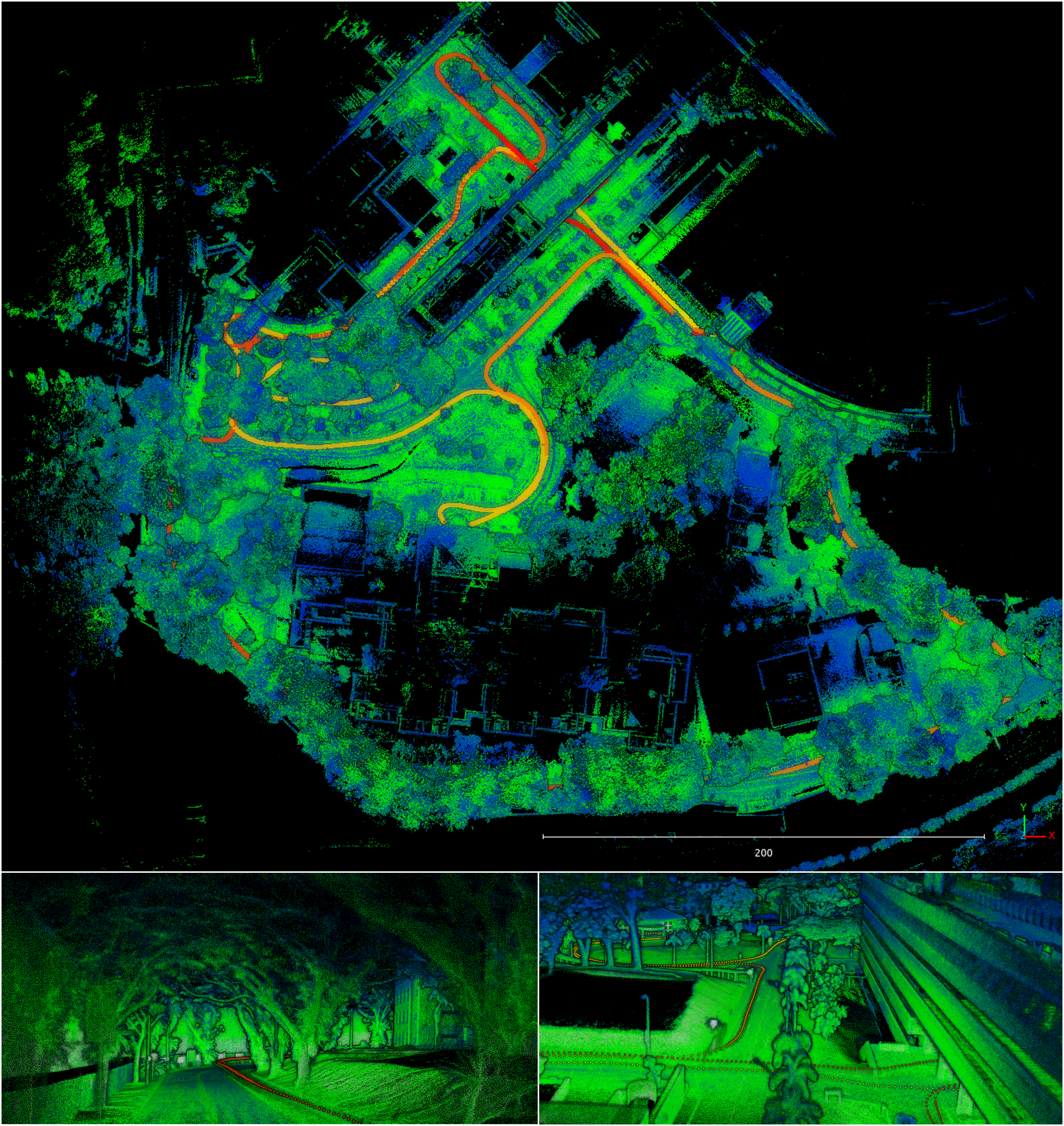}
	\end{overpic}
	\caption{The global map built by SLICT from one experiment. The two images at the bottom show the close-up views of the sections with dense vegetation and the junction areas mentioned in Fig. \ref{fig: mcd traj est}.}
	\label{fig: mcd exp nighttime 8}
\end{figure}

Since the change in elevation of the environment is significant (up to 15m between the highest and lowest points) the localization error is also significant, most visibly in the z dimension (Fig. \ref{fig: mcd error}). Compared to the Newer College dataset in Fig. \ref{fig: oxford global map exp 02 err}, we can see that the error in z direction is now almost doubled, which is the main contribution to the ATE in Tab. \ref{tab: ATE inhouse dataset}. In Fig. \ref{fig: mcd exp nighttime 8}, we present some views of the global map built by SLICT in sequence 2.

\begin{table}
\centering
\caption{ATE of SLICT and other methods on in-house dataset (unit [m]). The best results are in \tb{bold}, second best are \ul{underlined}. 'x' demotes a divergent experiment. LC denotes experiments with loop-closure and pose-graph optimization.}
\label{tab: ATE inhouse dataset}
\setlength{\tabcolsep}{5pt}
\begin{tabular}{ccccc|cc}
\hline\hline
\tb{Dataset}
&\tb{MARS}
&\tb{\begin{tabular}[c]{@{}c@{}}LIO-\\SAM\end{tabular}}
&\tb{\begin{tabular}[c]{@{}c@{}}FAST-\\LIO2\end{tabular}}
&\tb{\begin{tabular}[c]{@{}c@{}}SLICT\end{tabular}}
&\tb{\begin{tabular}[c]{@{}c@{}}LIO-\\SAM\\(LC)\end{tabular}}
&\tb{\begin{tabular}[c]{@{}c@{}}SLICT\\(LC)\end{tabular}}
\\ \hline

{seq\_01}
        &{x}			
        &{4.0784}       
        &\ul{2.0080}    
        &\tb{1.6929}    
        &\ul{1.2931}    
        &\tb{0.9815}\\  

{seq\_02}
        &{x}			
        &{3.7546}       
        &\ul{1.3623}    
        &\tb{1.0947}    
        &\ul{0.9685}    
        &\tb{0.7411}\\  
{seq\_03}
        &{x}			
        &{x}			
        &\ul{1.0926}    
        &\tb{0.8730}    
        &{x}			
        &\tb{0.8577}\\  

\hline\hline
\end{tabular}
\end{table}

\section{Conclusion} \label{sec: conclusion}

In this paper we have proposed a full-fledged multi-input Lidar-Inertial Odometry and Mapping system called SLICT. SLICT features a multi-scale global map of surfels that can be updated incrementally using the UFOMap framework. We also propose a method to associate lidar point to surfel (PTS), and a continuous-time MAP optimization of PTS and IMU-preintegration factors. We demonstrate competitive performance on public datasets even without loop closure.

To achieve a complete system, we add a simple yet effective loop closure mechanism and demonstrate its usefulness with our new datasets. We release the source code for the benefit of the community.

There remain many possibilities for extension of SLICT. For one, the PTS association still uses simple predicates, which can be made more efficient by some adaptive strategy. Moreover, the UFOMap framework allows one to integrate semantic information to the surfel, which has the potential to improve the association process. For the small drift, a basic loop closure mechanism sufficed, but more advanced methods can also be integrated in the future.


\balance
\bibliographystyle{IEEEtran}
\bibliography{references}

\begin{thebibliography}{10}
\providecommand{\url}[1]{#1}
\csname url@samestyle\endcsname
\providecommand{\newblock}{\relax}
\providecommand{\bibinfo}[2]{#2}
\providecommand{\BIBentrySTDinterwordspacing}{\spaceskip=0pt\relax}
\providecommand{\BIBentryALTinterwordstretchfactor}{4}
\providecommand{\BIBentryALTinterwordspacing}{\spaceskip=\fontdimen2\font plus
\BIBentryALTinterwordstretchfactor\fontdimen3\font minus
  \fontdimen4\font\relax}
\providecommand{\BIBforeignlanguage}[2]{{%
\expandafter\ifx\csname l@#1\endcsname\relax
\typeout{** WARNING: IEEEtran.bst: No hyphenation pattern has been}%
\typeout{** loaded for the language `#1'. Using the pattern for}%
\typeout{** the default language instead.}%
\else
\language=\csname l@#1\endcsname
\fi
#2}}
\providecommand{\BIBdecl}{\relax}
\BIBdecl

\bibitem{cadena2016past}
C.~Cadena, L.~Carlone, H.~Carrillo, Y.~Latif, D.~Scaramuzza, J.~Neira, I.~Reid,
  and J.~J. Leonard, ``Past, present, and future of simultaneous localization
  and mapping: Toward the robust-perception age,'' \emph{IEEE Transactions on
  Robotics}, vol.~32, no.~6, pp. 1309--1332, 2016.

\bibitem{zhang2014loam}
J.~Zhang and S.~Singh, ``Loam: Lidar odometry and mapping in real-time.'' in
  \emph{Robotics: Science and Systems}, vol.~2, no.~9, 2014.

\bibitem{shan2020liosam}
T.~Shan, B.~Englot, D.~Meyers, W.~Wang, C.~Ratti, and R.~Daniela, ``Lio-sam:
  Tightly-coupled lidar inertial odometry via smoothing and mapping,'' in
  \emph{IEEE/RSJ International Conference on Intelligent Robots and Systems
  (IROS)}.\hskip 1em plus 0.5em minus 0.4em\relax IEEE, 2020, pp. 5135--5142.

\bibitem{nguyen2021miliom}
T.-M. Nguyen, S.~Yuan, M.~Cao, Y.~Lyu, T.~H. Nguyen, and L.~Xie, ``Miliom:
  Tightly coupled multi-input lidar-inertia odometry and mapping,'' \emph{IEEE
  Robotics and Automation Letters}, vol.~6, no.~3, pp. 5573--5580, May 2021.

\bibitem{chen2021low}
P.~Chen, W.~Shi, S.~Bao, M.~Wang, W.~Fan, and H.~Xiang, ``Low-drift odometry,
  mapping and ground segmentation using a backpack lidar system,'' \emph{IEEE
  Robotics and Automation Letters}, 2021.

\bibitem{wang2021floam}
H.~Wang, C.~Wang, C.-L. Chen, and L.~Xie, ``F-loam: Fast lidar odometry and
  mapping,'' in \emph{2021 IEEE/RSJ International Conference on Intelligent
  Robots and Systems (IROS)}.\hskip 1em plus 0.5em minus 0.4em\relax IEEE,
  2021, pp. 4390--4396.

\bibitem{lv2021clins}
J.~Lv, K.~Hu, J.~Xu, Y.~Liu, X.~Ma, and X.~Zuo, ``Clins: Continuous-time
  trajectory estimation for lidar-inertial system,'' in \emph{2021 IEEE/RSJ
  International Conference on Intelligent Robots and Systems (IROS)}.\hskip 1em
  plus 0.5em minus 0.4em\relax IEEE, 2021, pp. 6657--6663.

\bibitem{xu2021fast}
W.~Xu and F.~Zhang, ``Fast-lio: A fast, robust lidar-inertial odometry package
  by tightly-coupled iterated kalman filter,'' \emph{IEEE Robotics and
  Automation Letters}, vol.~6, no.~2, pp. 3317--3324, 2021.

\bibitem{xu2022fast}
W.~Xu, Y.~Cai, D.~He, J.~Lin, and F.~Zhang, ``Fast-lio2: Fast direct
  lidar-inertial odometry,'' \emph{IEEE Transactions on Robotics}, 2022.

\bibitem{lin2021r2live}
J.~Lin, C.~Zheng, W.~Xu, and F.~Zhang, ``R 2 live: A robust, real-time,
  lidar-inertial-visual tightly-coupled state estimator and mapping,''
  \emph{IEEE Robotics and Automation Letters}, vol.~6, no.~4, pp. 7469--7476,
  2021.

\bibitem{lin2022r3live}
J.~Lin and F.~Zhang, ``R 3 live: A robust, real-time, rgb-colored,
  lidar-inertial-visual tightly-coupled state estimation and mapping package,''
  in \emph{2022 International Conference on Robotics and Automation
  (ICRA)}.\hskip 1em plus 0.5em minus 0.4em\relax IEEE, 2022, pp.
  10\,672--10\,678.

\bibitem{wisth2021unified}
D.~Wisth, M.~Camurri, S.~Das, and M.~Fallon, ``Unified multi-modal landmark
  tracking for tightly coupled lidar-visual-inertial odometry,'' \emph{IEEE
  Robotics and Automation Letters}, vol.~6, no.~2, pp. 1004--1011, 2021.

\bibitem{wisth2022vilens}
D.~Wisth, M.~Camurri, and M.~Fallon, ``Vilens: Visual, inertial, lidar, and leg
  odometry for all-terrain legged robots,'' \emph{IEEE Transactions on
  Robotics}, 2022.

\bibitem{quenzel2021real}
J.~Quenzel and S.~Behnke, ``Real-time multi-adaptive-resolution-surfel 6d lidar
  odometry using continuous-time trajectory optimization,'' in \emph{2021
  IEEE/RSJ International Conference on Intelligent Robots and Systems
  (IROS)}.\hskip 1em plus 0.5em minus 0.4em\relax IEEE, 2021, pp. 5499--5506.

\bibitem{bosse2012zebedee}
M.~Bosse, R.~Zlot, and P.~Flick, ``Zebedee: Design of a spring-mounted 3-d
  range sensor with application to mobile mapping,'' \emph{IEEE Transactions on
  Robotics}, vol.~28, no.~5, pp. 1104--1119, 2012.

\bibitem{bosse2009continuous}
M.~Bosse and R.~Zlot, ``Continuous 3d scan-matching with a spinning 2d laser,''
  in \emph{2009 IEEE International Conference on Robotics and
  Automation}.\hskip 1em plus 0.5em minus 0.4em\relax IEEE, 2009, pp.
  4312--4319.

\bibitem{duberg2020ufomap}
D.~Duberg and P.~Jensfelt, ``{UFOMap}: An efficient probabilistic {3D} mapping
  framework that embraces the unknown,'' \emph{IEEE Robotics and Automation
  Letters}, vol.~5, no.~4, pp. 6411--6418, 2020.

\bibitem{ye2019tightly}
H.~Ye, Y.~Chen, and M.~Liu, ``Tightly coupled 3d lidar inertial odometry and
  mapping,'' in \emph{2019 International Conference on Robotics and Automation
  (ICRA)}.\hskip 1em plus 0.5em minus 0.4em\relax IEEE, 2019, pp. 3144--3150.

\bibitem{welford1962note}
B.~Welford, ``Note on a method for calculating corrected sums of squares and
  products,'' \emph{Technometrics}, vol.~4, no.~3, pp. 419--420, 1962.

\bibitem{nguyen2021viral}
T.-M. Nguyen, M.~Cao, S.~Yuan, Y.~Lyu, T.~H. Nguyen, and L.~Xie,
  ``Viral-fusion: A visual-inertial-ranging-lidar sensor fusion approach,''
  \emph{IEEE Transactions on Robotics}, vol.~38, no.~2, pp. 958--977, 2022.

\bibitem{ceres-solver}
\BIBentryALTinterwordspacing
S.~Agarwal and K.~Mierle, ``Ceres solver: Tutorial \& reference.'' [Online].
  Available: \url{http://ceres-solver.org/}
\BIBentrySTDinterwordspacing

\bibitem{nguyen2021ntuviral}
T.-M. Nguyen, S.~Yuan, M.~Cao, Y.~Lyu, T.~H. Nguyen, and L.~Xie, ``Ntu viral: A
  visual-inertial-ranging-lidar dataset, from an aerial vehicle viewpoint,''
  \emph{The International Journal of Robotics Research}, vol.~41, no.~3, pp.
  270--280, 2022.

\bibitem{ramezani2020newer}
M.~Ramezani, Y.~Wang, M.~Camurri, D.~Wisth, M.~Mattamala, and M.~Fallon, ``The
  newer college dataset: Handheld lidar, inertial and vision with ground
  truth,'' in \emph{2020 IEEE/RSJ International Conference on Intelligent
  Robots and Systems (IROS)}, 2020, pp. 4353--4360.

\bibitem{gtsam}
\BIBentryALTinterwordspacing
F.~Dellaert, R.~Roberts, V.~Agrawal, A.~Cunningham, C.~Beall, D.-N. Ta,
  F.~Jiang, lucacarlone, nikai, J.~L. Blanco-Claraco, S.~Williams, ydjian,
  J.~Lambert, A.~Melim, Z.~Lv, A.~Krishnan, J.~Dong, G.~Chen, K.~Chande,
  balderdash devil, DiffDecisionTrees, S.~An, mpaluri, E.~P. Mendes, M.~Bosse,
  A.~Patel, A.~Baid, P.~Furgale, matthewbroadwaynavenio, and roderick koehle,
  ``borglab/gtsam,'' May 2022. [Online]. Available:
  \url{https://doi.org/10.5281/zenodo.5794541}
\BIBentrySTDinterwordspacing

\end{thebibliography}

\end{document}